\documentclass[letterpaper, 10 pt, conference]{ieeeconf}
\IEEEoverridecommandlockouts                              %
\usepackage[numbers]{natbib}
\usepackage{amsmath,amsfonts}
\usepackage{algorithmic}
\usepackage{algorithm}
\usepackage{array}
\usepackage{bm}
\usepackage{caption}
\usepackage{subcaption}
\usepackage{textcomp}
\usepackage{stfloats}
\usepackage{url}
\usepackage{verbatim}
\usepackage{graphicx}
\usepackage{mathabx}
\usepackage{xcolor}
\usepackage[draft=true, colorlinks=true, linkcolor=blue, citecolor=blue, urlcolor=blue, bookmarks=false]{hyperref}
\usepackage{xspace}
\renewcommand{\vec}{\bm}

\newcommand{\mat}{\mathbf}

\begin{document}

\title{\LARGE \bf Agile Continuous Jumping in Discontinuous Terrains}

\author{Yuxiang Yang$^1$, Guanya Shi$^2$, Changyi Lin$^2$, Xiangyun Meng$^1$, Rosario Scalise$^1$, Mateo Guaman Castro$^1$, \\Wenhao Yu$^3$, Tingnan Zhang$^3$, Ding Zhao$^2$, Jie Tan$^3$, and Byron Boots$^1$\\
$^1$University of Washington $^2$Carnegie Mellon University $^3$Google Deepmind
\thanks{$^1$ \texttt{\{yuxiangy, xiangyun, rosario, mateogc, bboots\}@ cs.washington.edu}}
\thanks{$^2$ \texttt{\{guanyas, changyil, dingzhao\}@andrew.cmu.edu}}
\thanks{$^3$ \texttt{\{magicmelon, tingnan, jietan\}@google.com}}
}

\maketitle

\begin{abstract}
We focus on agile, continuous, and terrain-adaptive jumping of quadrupedal robots in discontinuous terrains such as stairs and stepping stones.
Unlike single-step jumping, continuous jumping requires accurately executing highly dynamic motions over long horizons, which is challenging for existing approaches.
To accomplish this task, we design a hierarchical learning and control framework, which consists of a learned heightmap predictor for robust terrain perception, a reinforcement-learning-based centroidal-level motion policy for versatile and terrain-adaptive planning, and a low-level model-based leg controller for accurate motion tracking.
In addition, we minimize the sim-to-real gap by accurately modeling the hardware characteristics.
Our framework enables a Unitree Go1 robot to perform agile and continuous jumps on human-sized stairs and sparse stepping stones, for the first time to the best of our knowledge. In particular, the robot can cross two stair steps in each jump and completes a 3.5m long, 2.8m high, 14-step staircase in 4.5 seconds.
Moreover, the same policy outperforms baselines in various other parkour tasks, such as jumping over single horizontal or vertical discontinuities.
Experiment videos can be found at 
\url{https://yxyang.github.io/jumping\_cod/}.

\end{abstract}

\section{Introduction}

Achieving animal-level agility has long been a coveted goal in legged locomotion research. 
As a notable example, quadrupedal animals can traverse challenging terrains at high speeds with continuous jumping and precise foot placement.
Such jumping is particularly effective on discontinuous terrains like stairs and stepping stones, which is often non-traversable using the standard walking gait.
Inspired by this observation, researchers have made significant efforts to reproduce this agile jumping behavior in quadrupedal robots, including building high-performance robot platforms \cite{anymal, bledt2018cheetah, a1_robot, go1robot}, designing long-distance jumping controllers \cite{nguyen2022continuous, towr, gilroy2021autonomous, nguyen2019optimized}, and learning rapid hand-eye coordination \cite{cheng2023extreme, zhuang2023robot, barkour, hoeller2023anymal}.
However, achieving \emph{continuous}, \emph{long-distance}, and \emph{terrain-adaptive} jumping remains a challenging task.

While continuous, terrain-aware jumping has been studied in simulation \cite{peng2015dynamic, nguyen2021contact}, transferring this behavior to the real world remains difficult, and reflects many fundamental challenges in robot agility.
The first challenge is \emph{perception}. %
Despite recent progress, end-to-end learning-based methods \cite{agarwal2023legged, cheng2023extreme, zhuang2023robot, hoeller2023anymal} still face a large sim-to-real gap, especially for high-speed motions with significant camera oscillations.
The second challenge is \emph{motion accuracy}.
Unlike single-step jumping, continuous jumping requires the robot to link consecutive jumps with carefully planned body and foot motions, and accurately track these poses against unexpected perturbations.
Despite recent results in high-performance single-step jumping \cite{cheng2023extreme, zhuang2023robot, hoeller2023anymal}, most of these frameworks usually cannot maintain an accurate landing pose, making it difficult to extend from single-step to continuous jumping.
The last challenge is \emph{simulation fidelity}.
While domain randomization (DR) \cite{simtoreal, rma} can bridge small dynamics mismatches between simulation and the real world, it often falls short when the robot operates close to the hardware limit, which is common in agile, continuous jumping.
Because of these challenges, achieving high-speed jumping on real-world discontinuous terrains remains a difficult task.

\begin{figure}[t]
    \centering
    \includegraphics[width=.9\linewidth]{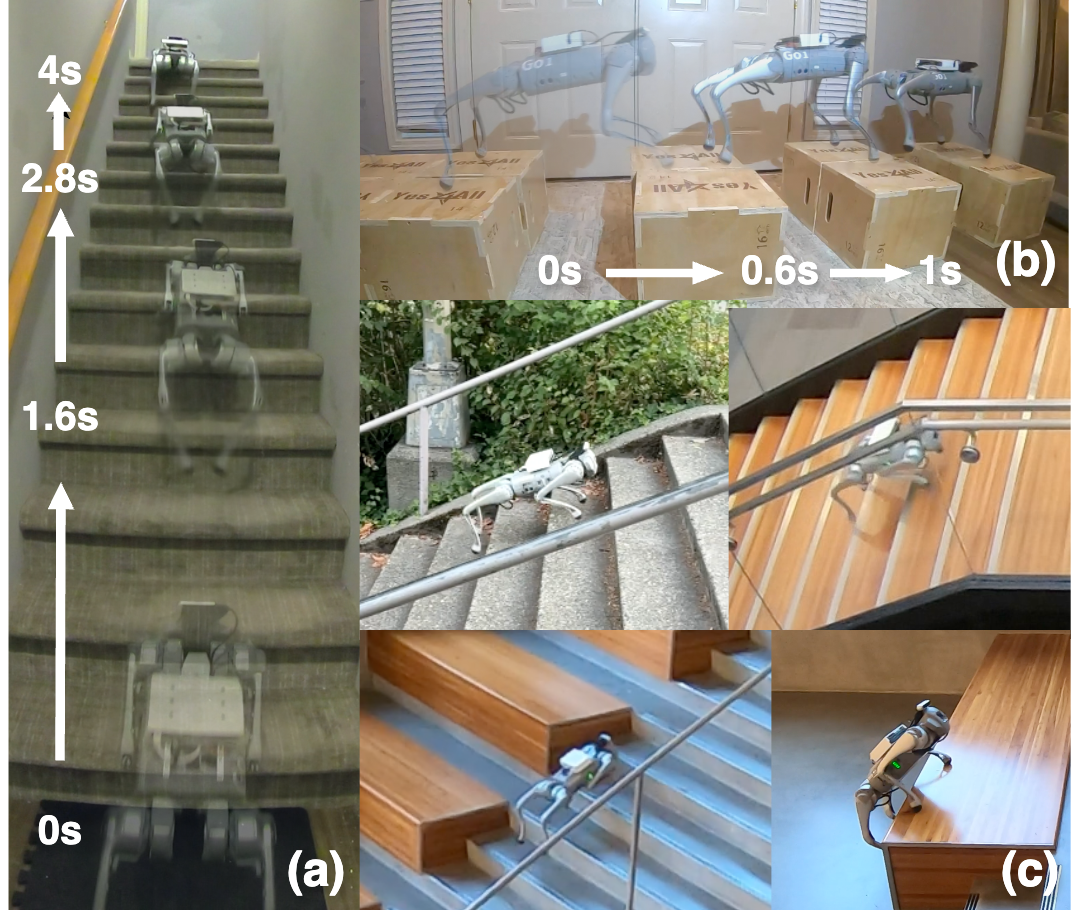}
    \caption{Our framework enables a quadrupedal robot to jump continuously over real-world stairs and steps.}
    \label{fig:teaser}
    \vspace{-1em}
\end{figure}

In this work, we present a hierarchical framework \cite{yang2023cajun} for agile, continuous, and terrain-aware jumping that addresses all these challenges.
Our framework consists of a learned heightmap predictor for terrain perception, a reinforcement learning (RL)-trained motion policy for motion planning, and a model-based leg controller for motion tracking.
Unlike end-to-end, pixel-to-action approaches \cite{cheng2023extreme, zhuang2023robot}, we choose the sagittal-plane heightmap as the intermediate representation, which improves the robustness and interpretability of terrain perception.
As the robot jumps forward, the heightmap predictor reconstructs this heightmap from onboard depth images.
Based on this heightmap, the motion policy plans body and foot motions, which is tracked by the leg controller.
To ensure consistent landing between jumps, we train the motion policy with an extra reward for accurate motion planning, and use a mixture of feedforward and feedback terms in the leg controller for robust motion tracking.
To improve simulation fidelity, we identify and reproduce key hardware characteristics in simulation, including camera delay and motor saturations.
Combining these efforts, our framework achieves animal-like agile continuous jumping on stairs and stepping stones for the first time. 

We conduct extensive real-robot experiments on a Unitree Go1 robot \cite{go1robot} to validate the effectiveness of our framework (Fig.~\ref{fig:teaser}).
With the continuous jumping gait, our robot jumps over a human-sized, 14-step staircase in less than 4.5 seconds, traversing 2 stair steps in each jump with an average horizontal speed of 0.8m/s and vertical speed of 0.6m/s.
This speed is more than 2 times faster than the standard walking gait (Table.~\ref{tab:jumping_performance}).
To the best of our knowledge, this is the first time any quadrupedal robot has achieved such high-speed animal-like traversal on similar staircases.
In addition to stair jumping, our framework also learns to jump continuously on horizontal stepping stones, and achieves state-of-the-art performance in single-step jumps up to 80cm ($2\times$ body length) horizontally and 60cm ($2.4\times$ body height) vertically \cite{zhuang2023robot, cheng2023extreme}.
We further conduct an ablation study to validate important design choices.

In summary, our contributions in this paper are as follows:

\begin{enumerate}
    \item We present a hierarchical framework for quadrupedal jumping with direct perception input.
    \item We develop a robust training pipeline that learns versatile jumping skills with minimal sim-to-real gap.
    \item Our framework achieves agile continuous jumps in real-world discontinuous terrains for the first time.
\end{enumerate}

\section{Related Work}

\subsection{Optimal Control for Dynamic Legged Locomotion}
Researchers have had a long history building optimal-control based controllers for dynamic quadrupedal locomotion \cite{mitcheetahmpc, mit_wbic, ethmpc, ethimpedance, villarreal2020mpc}.
By optimizing for control outputs at high frequency, these controllers can accurately track the reference trajectory, even for high-speed motions like running \cite{mitcheetahmpc}, galloping \cite{mit_wbic} or jumping \cite{mit_wbic,ding2019real,gehring2016practice}.
However, due to the constraint of real-time control, these frameworks are typically confined to a short plan horizon, and require additional effort in offline trajectory optimization \cite{gilroy2021autonomous, nguyen2019optimized, song2022optimal, towr}, dynamics model simplification \cite{nguyen2022continuous, pandala2022robust, li2024cafe}, and optimal control relaxation \cite{li2024cafe} to plan for complex, long-horizon jumps.
Another bottleneck of these frameworks is perception, which not only requires extra computation resource to process the perception signals, but also increases the complexity of the optimal control problem, making it more difficult to solve in real-time \cite{grandia2019feedback, qi2021perceptive}. 
Using manually-designed terrain perception and motion adaptation, \citet{park2021jumping} achieved bounding over fixed-shape hurdles on an MIT Cheetah 2 Robot.
More recently, by classifying step feasibility and formulating it into foot placement optimization, \citet{grandia2023perceptive} achieved low-speed walking on uneven stepping stones on an Anymal robot.
In this work, we use an optimal controller for robust low-level motion tracking, but introduce learning-based perception and motion planning for real-time replanning of complex jumping motions.

\subsection{Learning Perceptive Locomotion}
Recently, learning-based approaches emerge as a promising alternative for dynamic, terrain-adaptive quadrupedal locomotion \cite{margolis2022rapid, walk_these_ways, rma, rma_vision, zhuang2023robot, cheng2023extreme, twirl,He-RSS-24}.
The core idea is to construct an end-to-end policy that directly maps from perceptual and proprioceptive inputs to motor actions, and train the policy using reinforcement learning, usually in simulation \cite{makoviychuk2021isaac}.
These approaches have enabled legged robots to walk \cite{agarwal2023legged} and jump \cite{zhuang2023robot, cheng2023extreme, hoeller2023anymal, barkour} dynamically on challenging terrains such as mountain trails \cite{eth_hike} and stepping stones \cite{xie2020allsteps}.
Despite their success, these frameworks tend to be less robust on highly dynamic tasks, and face even more challenges on continuous jumping, which requires accurate motion planning and tracking.
Unlike these end-to-end approaches, we decompose our pipeline into perception, motion planning, and motion tracking, with interpretable intermediate outputs and robust real-world jumping performance.

\subsection{Hierarchical Frameworks}
Hierarchical frameworks offer a promising alternative that combines the benefit of learning-based and optimal-control-based controllers.
By combining a high-level learned policy with a low-level motor controller \cite{googlevisuallocomotion, laikagonvidia, glide, fast_and_efficient, yang2023cajun, yang2023continuous, bellegarda2020robust}, these frameworks can learn robust and generalizable locomotion behaviors with versatile gait selection \cite{laikagonvidia, fast_and_efficient}, foot placement \cite{glide, yang2023cajun, googlevisuallocomotion} and body motion planning \cite{glide, yang2023cajun}.
In our prior work \cite{yang2023cajun}, we addressed the computational bottleneck of hierarchical frameworks and achieved continuous forward jumping on flat terrains.
In this work, we extend the prior framework with perception, and achieve terrain-adaptive jumping in both forward and upward directions.

\section{Hierarchical Perceptive Jumping Framework}
\begin{figure*}[ht]
    \centering
    \includegraphics[width=\linewidth]{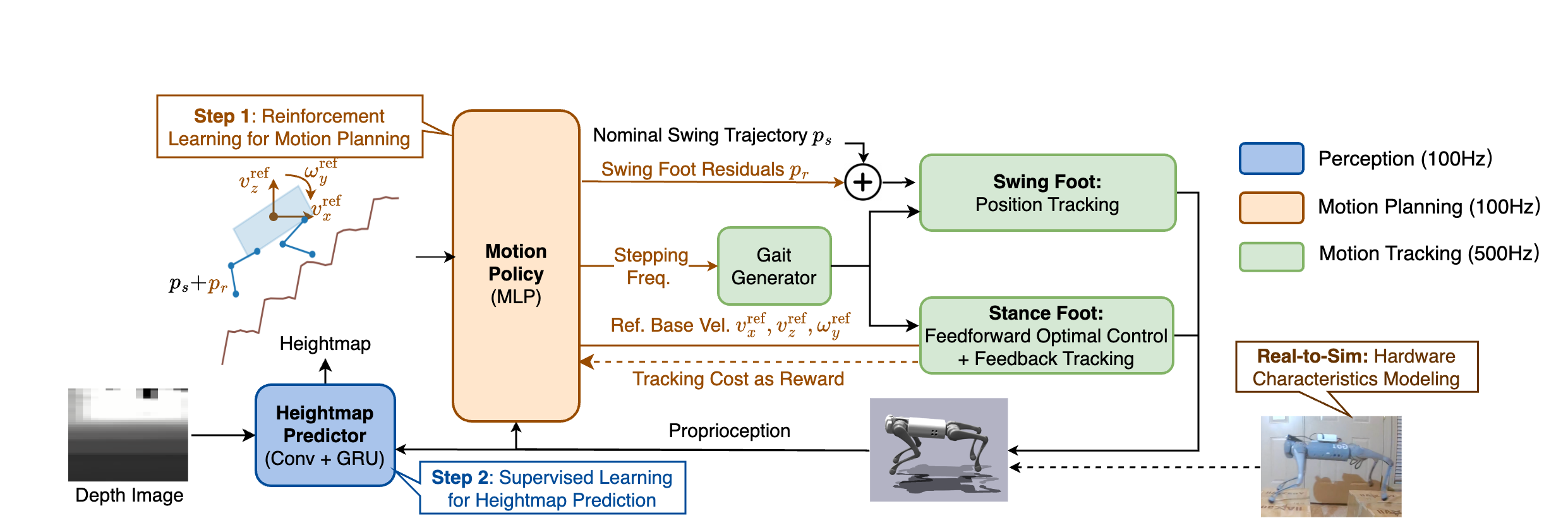}
    \vspace{-1.5em}
    \caption{\label{fig:block_diagram}Our hierarchical learning-control framework consists of a heightmap predictor, a motion policy, and a low-level leg controller. We use the heightmap as the intermediate representation for perception and motion planning (Section.~\ref{sec:heightmap}), train high-performance motion planning with reward to encourage accurate tracking (Section.~\ref{sec:motion_policy}), and combine a feedforward and a feedback controller for robust tracking of body orientations (Section.~\ref{sec:feedback}).
    In addition, we reduce the sim-to-real gap by accurately identifying key hardware characteristics and reproducing them in simulation (Section.~\ref{sec:real-to-sim}).}
    \vspace{-1.2em}
\end{figure*}

To achieve high-speed, terrain-aware, continuous jumping on the real robot, we design a hierarchical learning-control framework with separate modules for perception, motion planning, and motion tracking (Fig.~\ref{fig:block_diagram}).
At the perception level, a \emph{heightmap predictor} receives depth images from the onboard depth camera, and estimates a heightmap in the front-back axis of the robot.
Based on this heightmap, a \emph{motion policy} plans the reference robot motions, including the gait frequency $f$, the swing foot residuals $\vec{p}_r$, and the desired body velocity $v_x, v_z, v_{\theta}$.
This reference motion is then tracked by a low-level leg controller, which runs a gait generator to determine the desired contact state of each leg, and uses separate control strategies for swing and stance legs.
We run the motion policy and heightmap predictor at 100Hz for stable policy output, and run the leg controller at 500Hz for responsive torque control.

To train our framework, we first constructed a high-fidelity simulation environment with carefully identified hardware characteristics.
Due to the high computation cost in depth image rendering, we train our framework in two stages \cite{agarwal2023legged, cheng2023extreme, zhuang2023robot}.
In the first stage, we train the motion policy using reinforcement learning (RL), where the motion policy receives heightmaps that are directly sampled from the environment.
In the second stage, we train the heightmap predictor to estimate this heightmap from depth images.
Once the framework is trained, we deploy it to the real robot without additional fine-tuning.

\section{Heightmap-Centered Terrain Perception}
\label{sec:heightmap}
A robust and interpretable terrain understanding is critical for continuous jumping in noisy real-world environments. 
In our framework, we adopt the \emph{heightmap} as the intermediate representation between raw sensor data and motion planning. 
The heightmap predictor estimates this heightmap from depth images, while the motion policy uses it to plan body and foot motions.
This approach enhances the framework's interpretability and enables easy error isolation. 
Moreover, by randomizing the heightmap during training, we improve the motion policy's robustness against real-world uncertainties.

\subsection{Iterative Training for Heightmap Prediction}
We design the heightmap predictor as a 3-layer convolutional network followed by a 1-layer GRU, where the convolutional network extracts features from the depth images, and the GRU reconstructs the heightmap from past memories.
To ensure accurate heightmap reconstruction around the trajectories that the robot is more likely to visit, we train the heightmap predictor iteratively using supervised learning, similar to DAgger \cite{dagger}.
The training loop alternates between rolling out the trajectories using the latest heightmap predictor and the motion policy, and training the heightmap predictor on the trajectories collected.

\subsection{Robust Perception with Heightmap Randomization}
Due to camera occlusion and real-world sensor noises, it can be difficult to accurately reconstruct the ground-truth heightmap at all time.
To make the motion policy aware of this insufficiency, we randomly shift the observed heightmap during policy training ([-8cm, 8cm] horizontally, and [-5cm, 5cm] vertically).
This randomization improves the robustness of the motion policy and prevents it from overfitting to irrelevant details of the heightmap.
More importantly, it creates tolerance for inevitable heightmap reconstruction errors during real-world operation.

\section{Learning Versatile Motion Planning}
\label{sec:motion_policy}
To learn versatile jumping skills in challenging terrains, we train the motion policy using reinforcement learning, where the environment consists of diverse terrains in a curriculum of increasing difficulty.
In addition, to ensure the feasibility of planned motions, we include the motion tracking cost as part of the reward during policy training.

\subsection{Environment Setup}

In our environment, the state space includes robot proprioception, gait timing, and the randomized terrain heightmap.
The action space includes the stepping frequency $f$ for gait progress, the swing foot residual $\vec{p}_r$, and the reference base velocity $v_x^{\text{ref}}$, $v_z^{\text{ref}}, \omega_y^{\text{ref}}$.
We implement the environment and the low-level controller on GPU, and solve an approximated version of the optimal control problem \cite{yang2023cajun} to speed up training.
During real-world deployment, we solve the exact optimal control problem for extra robustness.

\subsection{Motion Tracking Cost as Policy Reward}
Note that not all motion references ($v_x^{\text{ref}}$, $v_z^{\text{ref}}, \omega_y^{\text{ref}}$) can be accurately tracked by the leg controller.
For example, when only the front leg is in contact, the robot cannot track a large $v_z^{\text{ref}}$ and a large $\omega_y^{\text{ref}}$ at the same time, as the prior requires jumping up with extended legs, and the later requires tilting down with retracted legs.
When facing such a conflicting objective, the leg controller typically computes an intermediate solution with large tracking errors.
Moreover, this solution can be highly sensitive to the current robot state, leading to large sim-to-real gaps and unpredictable real-world failures.
To alleviate this issue, we include the cost of the low-level optimal controler (Eq.~\ref{eq:qp_cost}) as part of the reward during policy training, so that the policy learns to plan feasible trajectories with small tracking errors.
The rest of the reward terms were adapted from our prior work \cite{yang2023cajun}, which encourages the robot to follow the contact schedule and jump over long distances.

\subsection{Terrain Curriculum for Skill Training}
\begin{figure}[t]
    \centering
    \includegraphics[trim=0em 0em 0em 0em,clip, width=1.\linewidth]{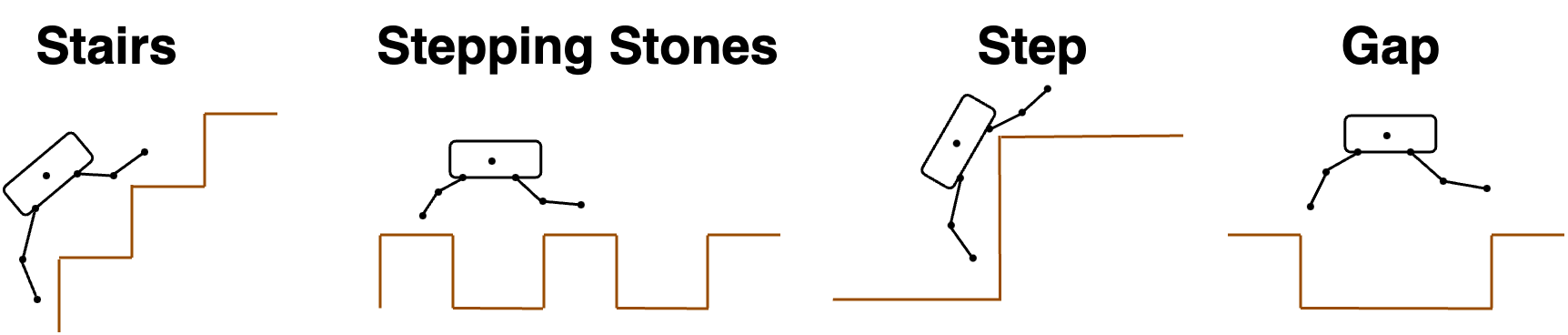}
    \vspace{-1.8em}
    \caption{The environment consists of 4 terrain types.}
    \label{fig:terrains}
\end{figure}

To learn continuous jumping over complex terrains, we design 4 terrain types in the training environment (Fig.~\ref{fig:terrains}) with single and multiple discontinuities.
Each terrain comes with a curriculum of increasing difficulty.
The robot starts randomly in the easiest terrain at the beginning of training, and advances to the next level after it learns to jump over a sufficiently long distance in the current level \cite{rudin2022learning}.

\section{Accurate Motion Tracking with Feedback}
\label{sec:feedback}
The leg controller tracks the motion references from the motion policy with separate control strategies for swing and stance legs.
The switch between swing and stance legs is modulated by a gait generator.
For accurate tracking of the body orientation changes, we combine a feedforward and a feedback controller for stance legs.

\subsection{Gait Generation and Swing Leg Control}

\begin{figure}[t]
    \centering
    \vspace{-1em}
    \includegraphics[trim=0em 0.2em 0em .7em, clip, width=1.\linewidth]{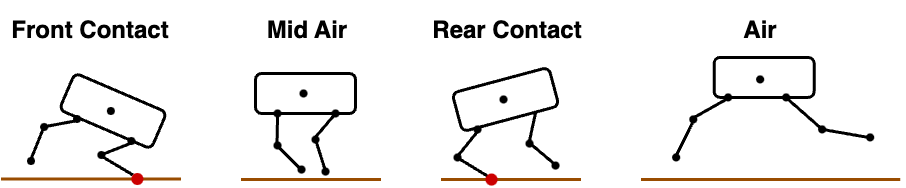}
    \vspace{-1.2em}
    \caption{We adopt the bounding gait with 4 contact modes. Red dot indicates foot contact.}    
    \label{fig:bounding_gait}
    \vspace{-2em}
\end{figure}

To maximize the jumping distance, the gait generator adopts the bounding gait with alternating foot contacts (Fig.~\ref{fig:bounding_gait}).
The gait generator tracks the leg's progress in each bounding cycle with a phase variable $\phi\in[0, 2\pi)$, where different values of $\phi$ correspond to different contact modes.
Given the stepping frequency $f$ from the motion policy, the gait generator advances the phase $\phi$ by $2\pi f$ and uses the newly computed phase to determine swing and stance legs.
The swing leg controller receives the reference foot positions as the sum of a nominal trajectory $\vec{p}_s$ from the Raibert heuristics \cite{raibertcontroller}, and a learned residual $\vec{p}_r$, converts the foot position to joint position using inverse kinematics (IK), and tracks this joint position using PD control.

\subsection{Feed-forward Optimal Control}
\begin{figure}[t]
    \centering
    \includegraphics[width=0.9\linewidth]{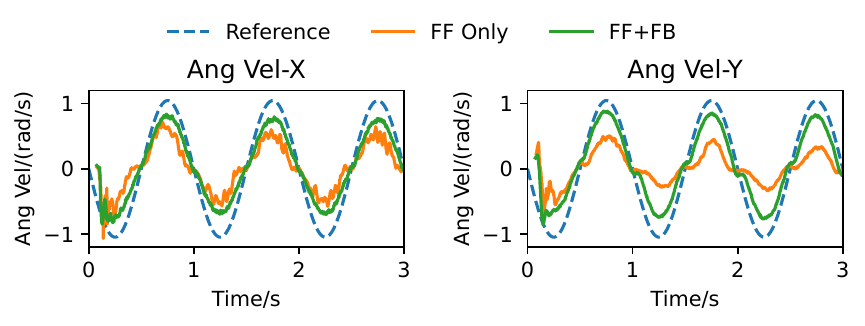}
    \vspace{-.8em}
    \caption{The feedback controller improves the tracking accuracy of body angular velocities.}
    \vspace{-1.5em}
    \label{fig:impedance}
\end{figure}

Given the reference base velocities $v_x^{\text{ref}}$, $v_z^{\text{ref}}$, $\omega_y^{\text{ref}}$, the feedforward controller computes a reference CoM acceleration $\vec{a}^{\text{ref}}\in\mathbb{R}^6$ via PD rule, and solves a quadratic program (QP) to find the ground reaction forces (GRFs) $\vec{f}\in\mathbb{R}^{12}$.
We set the objective to track the reference acceleration $\vec{a}^{\text{ref}}$ with an additional regularization term:
\begin{equation}
    \vspace{-0.2em}
    \min_{\vec{f}}\|\vec{a}-\vec{a}^{\text{ref}}\|_{\mat{U}} + \|\vec{f}\|_{\mat{V}}
    \label{eq:qp_cost}
    \vspace{-0.2em}
\end{equation}
where we compute the acceleration $\vec{a}$ via the linearized centroidal dynamics model~\cite{mitcheetahmpc}, and set $\mat{U}$ and $\mat{V}$ as diagonal weight matrices.
The constraints include the  contact schedule and friction cone constraints.
Once the GRFs are optimized, the controller then convert them into motor torques via Jacobian transpose: $\vec{\tau}=\mat{J}^{\top}\vec{f}$.

\subsection{Feedback Velocity Tracking}
\newcommand{\sref}{^{\text{ref}}}

While the feedforward (FF) controller can accurately track trajectories with upright body poses, we find that it cannot accurately track trajectories with large body orientation changes (Fig.~\ref{fig:impedance}) due to unmodeled details in robot orientation dynamics.
For accurate trajectory tracking under all body poses, we introduce a feedback (FB) controller \cite{mit_wbic, ethimpedance} to compensate for this tracking error.
This FB controller computes the reference foot \emph{velocities} from the reference body velocities, and tracks it using  joint PD control.
Note that the world frame position of a foot $\vec{p}_{\text{foot}}^{\text{W}}$ can be expressed in terms of its body frame position $\vec{p}_{\text{foot}}^{\text{B}}$ as:
\begin{equation}
    \vspace{-0.2em}\vec{p}_{\text{foot}}^{\text{W}}=\vec{p}_{\text{B}}^{\text{W}} + \mat{R}_{\text{B}}^{\text{W}}\vec{p}_{\text{foot}}^{\text{B}}
    \vspace{-0.2em}
\end{equation}
where $\vec{p}_{\text{B}}^{\text{W}}$ is the body position in the world frame, and $\mat{R}_{\text{B}}^{\text{W}}$ is the body rotation matrix.
Taking the derivative on both sides, we can express the world-frame foot velocity $\vec{v}_{\text{foot}}^{\text{W}}$ in terms of the body-frame foot velocity $\vec{v}_{\text{foot}}^{\text{B}}$:
\begin{equation}
    \vspace{-0.2em}
    \vec{v}_{\text{foot}}^{\text{W}}=\vec{v}_{\text{B}}^{\text{W}}+\vec{\omega}_{\text{B}}^{\text{W}}\times \vec{p}_{\text{foot}}^{\text{B}}+\mat{R}_{\text{B}}^{\text{W}}\vec{v}_{\text{foot}}^{\text{B}}
    \vspace{-0.2em}
\end{equation}
where $\vec{v}_{\text{B}}^{\text{W}}, \vec{\omega}_{\text{B}}^{\text{W}}$ is the body linear and angular velocity in the world frame.
Assuming static foot contacts $\left(\vec{v}_{\text{foot}}^{\text{W}}=0\right)$ and that the body is moving at reference velocities, we can express the reference body-frame foot velocities $\vec{v}_{\text{foot}}^{\text{B}}$ in terms of the reference body velocities.
\begin{equation}
    \vspace{-0.2em}
    \vec{v}_{\text{foot}}^{\text{B}}=-\left(\mat{R}_{\text{B}}^{\text{W}}\right)^{-1}(\vec{v}\sref + \vec{\omega}\sref\times\vec{p}_{\text{foot}}^{\text{B}})
    \vspace{-0.2em}
\end{equation}

We then convert the reference foot velocities to the reference joint velocities via Jacobian inverse $\dot{\vec{q}}\sref = \mat{J}^{-1}\vec{v}_{\text{foot}}^{\text{B}}$ and compute the feedback torques as $\vec{\tau}_{\text{fb}}=\vec{k}_d^{\text{fb}}(\dot{\vec{q}}\sref  - \dot{\vec{q}})$.
We set $\vec{k}_d^{\text{fb}}=1$ to provide effective feedback while not causing significant interference with the feedforward controller.

\newcommand{\pref}{\vec{p}^{\text{ref}}}
\newcommand{\vref}{\vec{v}^{\text{ref}}}
\newcommand{\aref}{\vec{a}^{\text{ref}}}

\section{High-fidelity Simulation with Real-to-Sim}
\label{sec:real-to-sim}
Sim-to-real is particularly challenging for highly dynamic jumping tasks.
While domain randomization (DR) can improve the robustness of the policy, randomization of standard simulation parameters may not capture the unique behavior of real hardware, especially for highly dynamic tasks that operate close to hardware limit.
In this work, we design individual experiments to carefully measure these characteristics, and reproduce them to improve simulation fidelity.

\begin{figure}
    \centering
    \begin{subfigure}{.48\linewidth}
        \centering
        \includegraphics[width=\linewidth]{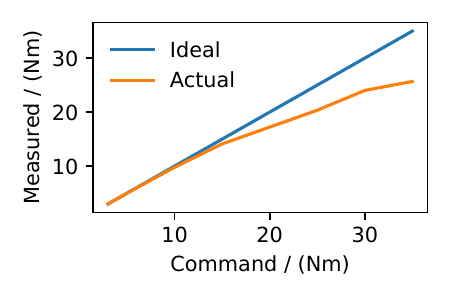}
        \vspace{-2em}
        \caption{Torque Saturation.}
        \label{fig:torque_saturation}
    \end{subfigure}
    \begin{subfigure}{.48\linewidth}
        \centering
        \includegraphics[width=\linewidth]{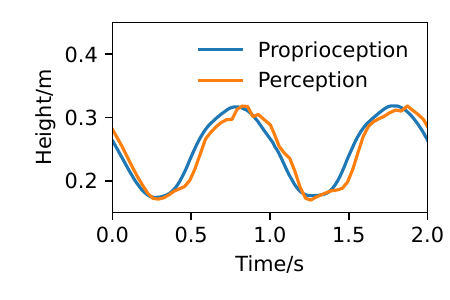}
        \vspace{-2em}
        \caption{Camera Latency.}
        \label{fig:perception_delay}
    \end{subfigure}
    \vspace{-.4em}
    \caption{We conduct individual studies to measure the motor torque saturations (Fig.~\ref{fig:torque_saturation}) and camera latency (Fig.~\ref{fig:perception_delay}).}
    \vspace{-2em}
\end{figure}

\paragraph{Measuring Motor Saturation}
Jumping over long distances requires the robot to fully utilize its motor torque outputs to generate the necessary foot forces.
Assuming a linear torque-current relationship, standard DC motors track a torque command by tracking the current output.
However, at high torque commands, the torque-current relationship tend to saturate, leading to insufficient torque output at the desired current \cite{simtoreal}.
To accurately capture this torque saturation, we measure the motor torque output at different torque commands using a dynamometer (Fig.~\ref{fig:torque_saturation}).
We then reproduce this relationship in simulation using a piecewise linear function, and convert all torque commands before stepping the simulation \cite{simtoreal}.

\paragraph{Measuring Camera Latency}
To rapidly respond to terrain changes, its critical for the robot to understand any latency in terrain perception.
To measure this camera latency, we execute sinusoidal body height motions on the real robot, and compare the body height estimated from joint angles using forward kinematics with the body height estimated from the received depth images (Fig.\ref{fig:perception_delay}). 
The perception-based height estimation shows a consistent delay of about 50ms, or 5 environment steps.
We subsequently implement this delay in all depth image rendering in simulation.

\section{Results and Analysis}
\begin{table*}[ht]
    \centering
    \vspace{1em}
    \begin{tabular}{c|c|c|c|c|c|c}
        \hline
        &Robot&Style & Stairs & Stepping Stones &  Step & Gap  \\
        &&&(Max Speed) & (Max Speed) & (Max Height) & (Max Width)\\\hline
        Egocentric-Vision\cite{agarwal2023legged}& A1 & Walking&0.46m/s&0.36m/s&0.26m (1$\times$)&0.17m (0.43$\times$)\\
        Robot Parkour \cite{zhuang2023robot} &A1 & Jumping &-&-& 0.4m (1.6$\times$)&0.6m (1.5$\times$) \\
        Extreme Parkour \cite{cheng2023extreme} &A1& Jumping&- & - &0.5m (2$\times$)&\textbf{0.8m (2$\times$)}\\
        \hline
        Ours&Go1& Jumping&\textbf{1m/s} & \textbf{1.4m/s}  &\textbf{0.6m (2.4$\times$)}&\textbf{0.8m (2$\times$)}\\
        \hline
    \end{tabular}
    \vspace{-0.4em}
    \caption{Performance comparison between our framework and baselines. Multiplier ($\times$) shows distance relative to body dimensions (length or height).}
    \label{tab:jumping_performance}
    \vspace{-2.6em}
\end{table*}

We design experiments to validate the performance of our framework in agile jumping over discontinuous terrains. More specifically, we aim to answer the following questions:

\begin{enumerate}
    \item Can our framework jump over challenging discontinuous terrains at high speed in the real world?
    \item How does the jumping performance of our framework compare with existing works?
    \item Can our framework perform terrain-aware body and foot-step planning?
    \item What are the important design choices to facilitate successful sim-to-real transfer?
\end{enumerate}
\subsection{Experimental Setup}

We test our framework on the Unitree Go1 robot \cite{go1robot}, where we mounted a Intel Realsense D435i camera to capture depth images, and stream the images to a Mac Mini computer for policy inference. 
We build the simulation environment in IsaacGym \cite{makoviychuk2021isaac}, and implemented the low-level leg controller in PyTorch \cite{pytorch}.
We train the motion policy for 8000 gradient steps using Proximal Policy Optimization (PPO) \cite{ppo}, which takes about 7 hours on a Nvidia RTX 4090 GPU.
We train the heightmap predictor for 30 DAgger steps where each step collects 5000 state-action pairs.
Training the heightmap predictor takes about 1 hour on the same computer.

\subsection{Continuous Jumping over Discontinuous Terrains}

We deploy the trained framework in two real-world discontinuous terrains (Fig.~\ref{fig:teaser} (a) (b)), including a staircase and a stepping stone environment.
The robot carefully coordinates its body and foot motion and traverses through both terrains at high speeds.
In the first case, the robot completes the 14-step staircase in 4.5 seconds, with an average horizontal speed of 0.8m/s and vertical speed of 0.6m/s.
A closer look at the jumping behavior reveals that the robot crosses the entire 14-step stair cases in 8 jumps, with 2 stair steps per jump most of the time.
In the second case, we test the robot on a stepping stone environment.%
The robot jumps across one gap each time, and traverses the entire terrain in less than 2 seconds.
To the best of our knowledge, this is the first time a quadrupedal robot has demonstrated such agile continuous jumping on these challenging terrains.

We compare the terrain traversal performance of our framework with a few baselines in the first two columns of Table.~\ref{tab:jumping_performance}.
Jumping significantly improves the robot's traversal speed in these terrains, resulting in more than 2 times speedup compared to the walking gait \cite{agarwal2023legged}.
In addition, \emph{continuous} jumping is important for these terrains with repeated discontinuities.
While prior works \cite{cheng2023extreme, zhuang2023robot} have demonstrated high-performance single-step jumps, these results do not generalize to continuous jumps, which requires accurate and robust motion control over long horizons.

To test the generalizability of our framework, we deploy our framework on 4 real-world terrains of different geometries and materials (Fig.~\ref{fig:teaser}(c)).
Our framework performs well on 3 of them with a success rate of 100\%.
The only exception is the staircase on the top-right of Fig.~\ref{fig:teaser}(c), where the robot achieved a success rate of 20\%.
This staircase features a unique geometry with disconnected planked steps, which is not seen during training.
Therefore, both the motion policy and the heightmap predictor had difficulty recognizing and reacting to this stair case.

\subsection{High-performance Jumping over a Single Discontinuity}

\begin{figure}[t]
    \centering
    \includegraphics[width=.9\linewidth]{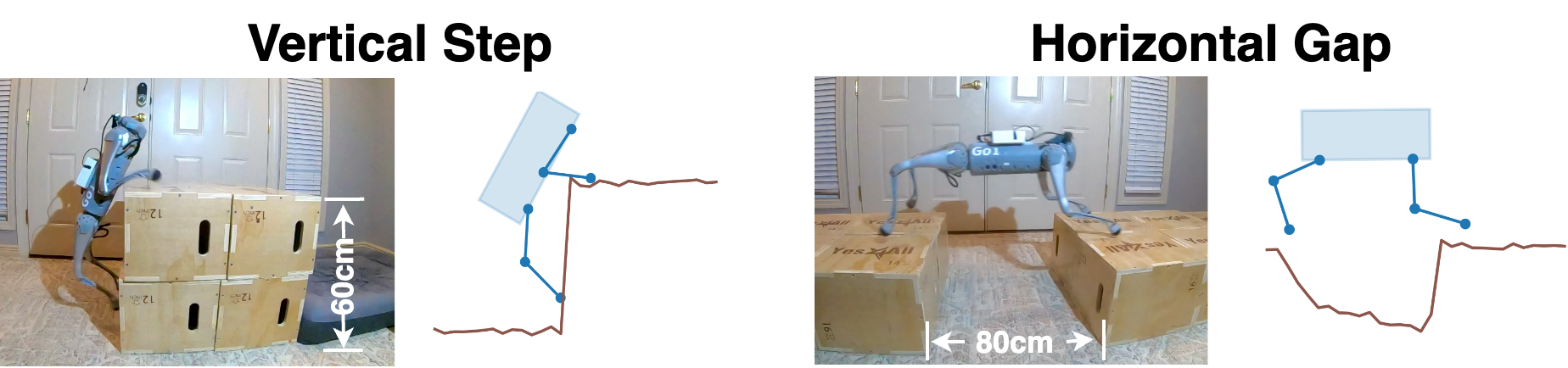}    
    \caption{\label{fig:single_discontinuity}Key frames and the estimated heightmap of the robot jumping over a 60cm step (left) and a 80cm gap (right).}
    \vspace{-0.5em}
\end{figure}

To further validate the capability of our framework, we test our framework in manually constructed terrains with a single discontinuity, including a vertical step or a horizontal gap (Fig.~\ref{fig:single_discontinuity}).
Our framework handles these terrains well, and can cross horizontal gaps up to 80cm and climb vertical steps up to 60cm.
We further visualize the estimated heightmap and the robot pose in Fig.~\ref{fig:single_discontinuity}.
Using only proprioception and ego-centric depth image, the heightmap predictor accurately estimates the terrain shape with small contact errors.
This jumping performance matches or even exceeds the performance of prior works on similar-sized robots (Table.~\ref{tab:jumping_performance}), even though our hardware platform, the Go1 Robot \cite{go1robot}, has weaker motor outputs compared to the prior platform (A1).

\begin{figure}[t]
    \centering
    \vspace{-0.5em}
    \includegraphics[width=.9\linewidth]{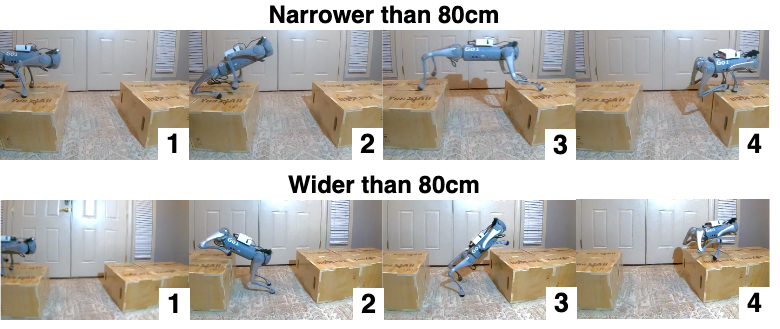}
    \caption{The motion policy plans different foot placements for different gap widths. \textbf{Top}: For narrower gaps, the robot jumps over the gap directly. \textbf{Bottom}: for wider gaps, the robot lands on the ground and jumps over the gap again.}
    \label{fig:adaptive_gap}
    \vspace{-1.5em}
\end{figure}

\begin{figure}[t]
    \centering
    \includegraphics[trim=0em 0em 0em 0em,clip,width=.9\linewidth]{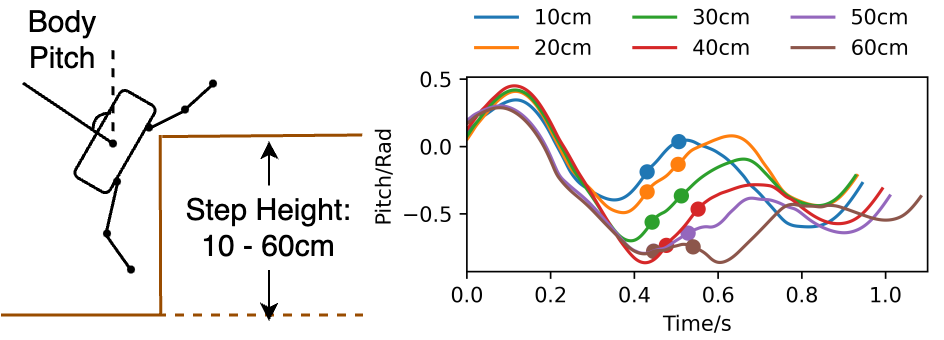}
    \vspace{-.7em}
    \caption{The motion policy plans different body pitch trajectories when jumping over different step heights. The markers indicates the air phase of the bounding gait (Fig.~\ref{fig:bounding_gait}).}
    \label{fig:adaptive_step}
    \vspace{-.8em}
\end{figure}

\subsection{Emergent Foot and Body Motion Planning}
As the core of our framework, the motion policy learns versatile body and foot motion planning based on perceived terrain. For example, the motion policy plans different foot placements when jumping over gaps of different sizes (Fig.~\ref{fig:adaptive_gap}), from single-step jumping for narrower gaps to two-step jumping with intermediate landing for wider gaps.
The motion policy also plans different body pitch trajectories when jumping on steps of different heights (Fig.~\ref{fig:adaptive_step}), and prefers larger pitch angles for higher steps.

\subsection{Ablation Study}
\begin{figure}[t]
    \centering
    \includegraphics[trim=0em 0em 0em 1.5em, clip, width=\linewidth]{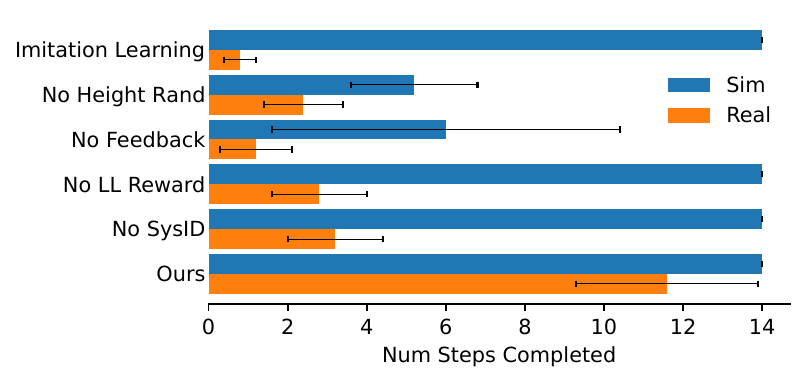}
    \vspace{-2em}
    \caption{The number of steps before failure (over 5 trials) in simulation and in the real world on a 14-step staircase.}
    \vspace{-1.5em}
    \label{fig:ablation_real}
\end{figure}

To validate the design choices of our framework, we perform an ablation study, and test the simulation and real-world performance of our frameworks and the baselines on a 14-step stair case (Fig.~\ref{fig:ablation_real}).
We run each baseline 5 times, and report the number of steps reached before failure.
Most baselines perform well in simulation, and complete the entire 14-step staircase.
One exception is the baseline where the motion policy is trained without heightmap randomization (No Height Rand).
While this baseline can jump well with the ground-truth heightmap, it cannot keep a similar performance even with small errors in heightmap estimation.
Another exception is the baseline where the stance leg does not use the feedback controller (No Feedback).
We find that the robot cannot track accurate body orientation changes, and cannot perform well on vertical jumps, which is consistent with prior observations (Fig.~\ref{fig:impedance}).

While the rest of the baselines completed the entire 14-step staircases in simulation, their performances drop significantly when transferred to the real world.
Without the heightmap as the intermediate terrain representation, the imitation learning baseline, which directly imitates the motion policy's actions from depth images, cannot jump more than 2 staircases in the real world due to perception noise.
Without the low-level motion tracking reward, the No LL Reward baseline manages to jump for a few steps in the real world, but eventually falls over.
Lastly, without knowledge of the sensor and actuator characteristics, the "No SysID" baseline jumps very fast in the real world and crashes into the stair edges, likely due to its unawareness of motor saturation.

\section{Conclusion}
In this work, we present a hierarchical framework for agile continuous jumping in discontinuous terrains.
Using our framework, a quadrupedal robot achieves animal-like high-speed jumping on stairs and stepping stones for the first time, and outperforms baselines in various other parkour tasks.
One limitation of our framework is that the robot motion is limited to the sagittal plane, and does not support sideways or turning motions.
Another limitation is that our framework follows a fixed directional command, and does not strategically plan a path to reach the destination.
In future work, we plan to address these limitations by expanding the scope of the current training pipeline, and achieve versatile, goal-oriented agile locomotion in complex terrains.

\newpage
\bibliographystyle{IEEEtranN}
\bibliography{reference}  %

\begin{thebibliography}{51}
\providecommand{\natexlab}[1]{#1}
\providecommand{\url}[1]{#1}
\csname url@samestyle\endcsname
\providecommand{\newblock}{\relax}
\providecommand{\bibinfo}[2]{#2}
\providecommand{\BIBentrySTDinterwordspacing}{\spaceskip=0pt\relax}
\providecommand{\BIBentryALTinterwordstretchfactor}{4}
\providecommand{\BIBentryALTinterwordspacing}{\spaceskip=\fontdimen2\font plus
\BIBentryALTinterwordstretchfactor\fontdimen3\font minus \fontdimen4\font\relax}
\providecommand{\BIBforeignlanguage}[2]{{%
\expandafter\ifx\csname l@#1\endcsname\relax
\typeout{** WARNING: IEEEtranN.bst: No hyphenation pattern has been}%
\typeout{** loaded for the language `#1'. Using the pattern for}%
\typeout{** the default language instead.}%
\else
\language=\csname l@#1\endcsname
\fi
#2}}
\providecommand{\BIBdecl}{\relax}
\BIBdecl

\bibitem[ANYbotics(2020)]{anymal}
\BIBentryALTinterwordspacing
ANYbotics, ``Anymal: Autonomous legged robot,'' Product Information, 2020. [Online]. Available: \url{https://www.anybotics.com/anymal}
\BIBentrySTDinterwordspacing

\bibitem[Bledt et~al.(2018)Bledt, Powell, Katz, Di~Carlo, Wensing, and Kim]{bledt2018cheetah}
G.~Bledt, M.~J. Powell, B.~Katz, J.~Di~Carlo, P.~M. Wensing, and S.~Kim, ``Mit cheetah 3: Design and control of a robust, dynamic quadruped robot,'' in \emph{2018 IEEE/RSJ International Conference on Intelligent Robots and Systems (IROS)}.\hskip 1em plus 0.5em minus 0.4em\relax IEEE, 2018, pp. 2245--2252.

\bibitem[a1_(2020)]{a1_robot}
\BIBentryALTinterwordspacing
``\BIBforeignlanguage{en}{A1 {Website}},'' 2020. [Online]. Available: \url{https://www.unitree.com/products/a1/}
\BIBentrySTDinterwordspacing

\bibitem[go1(2021)]{go1robot}
\BIBentryALTinterwordspacing
``\BIBforeignlanguage{en}{Go1 {Website}},'' 2021. [Online]. Available: \url{https://www.unitree.com/products/go1/}
\BIBentrySTDinterwordspacing

\bibitem[Nguyen et~al.(2022)Nguyen, Bao, and Nguyen]{nguyen2022continuous}
C.~Nguyen, L.~Bao, and Q.~Nguyen, ``Continuous jumping for legged robots on stepping stones via trajectory optimization and model predictive control,'' in \emph{2022 IEEE 61st Conference on Decision and Control (CDC)}.\hskip 1em plus 0.5em minus 0.4em\relax IEEE, 2022, pp. 93--99.

\bibitem[Winkler et~al.(2018)Winkler, Bellicoso, Hutter, and Buchli]{towr}
A.~W. Winkler, C.~D. Bellicoso, M.~Hutter, and J.~Buchli, ``Gait and trajectory optimization for legged systems through phase-based end-effector parameterization,'' \emph{IEEE Robotics and Automation Letters}, vol.~3, no.~3, pp. 1560--1567, 2018.

\bibitem[Gilroy et~al.(2021)Gilroy, Lau, Yang, Izaguirre, Biermayer, Xiao, Sun, Agrawal, Zeng, Li, et~al.]{gilroy2021autonomous}
S.~Gilroy, D.~Lau, L.~Yang, E.~Izaguirre, K.~Biermayer, A.~Xiao, M.~Sun, A.~Agrawal, J.~Zeng, Z.~Li \emph{et~al.}, ``Autonomous navigation for quadrupedal robots with optimized jumping through constrained obstacles,'' in \emph{2021 IEEE 17th International Conference on Automation Science and Engineering (CASE)}.\hskip 1em plus 0.5em minus 0.4em\relax IEEE, 2021, pp. 2132--2139.

\bibitem[Nguyen et~al.(2019)Nguyen, Powell, Katz, Di~Carlo, and Kim]{nguyen2019optimized}
Q.~Nguyen, M.~J. Powell, B.~Katz, J.~Di~Carlo, and S.~Kim, ``Optimized jumping on the mit cheetah 3 robot,'' in \emph{2019 International Conference on Robotics and Automation (ICRA)}.\hskip 1em plus 0.5em minus 0.4em\relax IEEE, 2019, pp. 7448--7454.

\bibitem[Cheng et~al.(2023)Cheng, Shi, Agarwal, and Pathak]{cheng2023extreme}
X.~Cheng, K.~Shi, A.~Agarwal, and D.~Pathak, ``Extreme parkour with legged robots,'' \emph{arXiv preprint arXiv:2309.14341}, 2023.

\bibitem[Zhuang et~al.(2023)Zhuang, Fu, Wang, Atkeson, Schwertfeger, Finn, and Zhao]{zhuang2023robot}
Z.~Zhuang, Z.~Fu, J.~Wang, C.~G. Atkeson, S.~Schwertfeger, C.~Finn, and H.~Zhao, ``Robot parkour learning,'' in \emph{CoRL 2023}, 2023.

\bibitem[Caluwaerts et~al.(2023)Caluwaerts, Iscen, Kew, Yu, Zhang, Freeman, Lee, Lee, Saliceti, Zhuang, et~al.]{barkour}
K.~Caluwaerts, A.~Iscen, J.~C. Kew, W.~Yu, T.~Zhang, D.~Freeman, K.-H. Lee, L.~Lee, S.~Saliceti, V.~Zhuang \emph{et~al.}, ``Barkour: Benchmarking animal-level agility with quadruped robots,'' \emph{arXiv preprint arXiv:2305.14654}, 2023.

\bibitem[Hoeller et~al.(2023)Hoeller, Rudin, Sako, and Hutter]{hoeller2023anymal}
D.~Hoeller, N.~Rudin, D.~Sako, and M.~Hutter, ``Anymal parkour: Learning agile navigation for quadrupedal robots,'' \emph{arXiv preprint arXiv:2306.14874}, 2023.

\bibitem[Peng et~al.(2015)Peng, Berseth, and Van~de Panne]{peng2015dynamic}
X.~B. Peng, G.~Berseth, and M.~Van~de Panne, ``Dynamic terrain traversal skills using reinforcement learning,'' \emph{ACM Transactions on Graphics (TOG)}, vol.~34, no.~4, pp. 1--11, 2015.

\bibitem[Nguyen and Nguyen(2021)]{nguyen2021contact}
C.~Nguyen and Q.~Nguyen, ``Contact-timing and trajectory optimization for 3d jumping on quadruped robots. arxiv. 10.48550,'' \emph{arXiv preprint ARXIV.2110.06764}, 2021.

\bibitem[Agarwal et~al.(2023)Agarwal, Kumar, Malik, and Pathak]{agarwal2023legged}
A.~Agarwal, A.~Kumar, J.~Malik, and D.~Pathak, ``Legged locomotion in challenging terrains using egocentric vision,'' in \emph{Conference on Robot Learning}.\hskip 1em plus 0.5em minus 0.4em\relax PMLR, 2023, pp. 403--415.

\bibitem[Tan et~al.(2018)Tan, Zhang, Coumans, Iscen, Bai, Hafner, Bohez, and Vanhoucke]{simtoreal}
J.~Tan, T.~Zhang, E.~Coumans, A.~Iscen, Y.~Bai, D.~Hafner, S.~Bohez, and V.~Vanhoucke, ``Sim-to-real: Learning agile locomotion for quadruped robots,'' in \emph{Proceedings of Robotics: Science and Systems}, Pittsburgh, Pennsylvania, June 2018.

\bibitem[Kumar et~al.(2021)Kumar, Fu, Pathak, and Malik]{rma}
A.~Kumar, Z.~Fu, D.~Pathak, and J.~Malik, ``Rma: Rapid motor adaptation for legged robots,'' \emph{Robotics: Science and Systems}, 2021.

\bibitem[Yang et~al.(2023{\natexlab{a}})Yang, Shi, Meng, Yu, Zhang, Tan, and Boots]{yang2023cajun}
Y.~Yang, G.~Shi, X.~Meng, W.~Yu, T.~Zhang, J.~Tan, and B.~Boots, ``Cajun: Continuous adaptive jumping using a learned centroidal controller,'' \emph{arXiv preprint arXiv:2306.09557}, 2023.

\bibitem[{Di Carlo} et~al.(2018){Di Carlo}, {Wensing}, {Katz}, {Bledt}, and {Kim}]{mitcheetahmpc}
J.~{Di Carlo}, P.~M. {Wensing}, B.~{Katz}, G.~{Bledt}, and S.~{Kim}, ``Dynamic locomotion in the mit cheetah 3 through convex model-predictive control,'' in \emph{2018 IEEE/RSJ International Conference on Intelligent Robots and Systems (IROS)}, 2018, pp. 1--9.

\bibitem[Kim et~al.(2019)Kim, Di~Carlo, Katz, Bledt, and Kim]{mit_wbic}
D.~Kim, J.~Di~Carlo, B.~Katz, G.~Bledt, and S.~Kim, ``Highly dynamic quadruped locomotion via whole-body impulse control and model predictive control,'' \emph{arXiv preprint arXiv:1909.06586}, 2019.

\bibitem[Horvat et~al.(2017)Horvat, Melo, and Ijspeert]{ethmpc}
T.~Horvat, K.~Melo, and A.~J. Ijspeert, ``Model predictive control based framework for com control of a quadruped robot,'' in \emph{2017 IEEE/RSJ International Conference on Intelligent Robots and Systems (IROS)}.\hskip 1em plus 0.5em minus 0.4em\relax IEEE, 2017, pp. 3372--3378.

\bibitem[Jenelten et~al.(2019)Jenelten, Hwangbo, Tresoldi, Bellicoso, and Hutter]{ethimpedance}
F.~Jenelten, J.~Hwangbo, F.~Tresoldi, C.~D. Bellicoso, and M.~Hutter, ``Dynamic locomotion on slippery ground,'' \emph{IEEE Robotics and Automation Letters}, vol.~4, no.~4, pp. 4170--4176, 2019.

\bibitem[Villarreal et~al.(2020)Villarreal, Barasuol, Wensing, Caldwell, and Semini]{villarreal2020mpc}
O.~Villarreal, V.~Barasuol, P.~M. Wensing, D.~G. Caldwell, and C.~Semini, ``Mpc-based controller with terrain insight for dynamic legged locomotion,'' in \emph{2020 IEEE International Conference on Robotics and Automation (ICRA)}.\hskip 1em plus 0.5em minus 0.4em\relax IEEE, 2020, pp. 2436--2442.

\bibitem[Ding et~al.(2019)Ding, Pandala, and Park]{ding2019real}
Y.~Ding, A.~Pandala, and H.-W. Park, ``Real-time model predictive control for versatile dynamic motions in quadrupedal robots,'' in \emph{2019 International Conference on Robotics and Automation (ICRA)}.\hskip 1em plus 0.5em minus 0.4em\relax IEEE, 2019, pp. 8484--8490.

\bibitem[Gehring et~al.(2016)Gehring, Coros, Hutter, Bellicoso, Heijnen, Diethelm, Bloesch, Fankhauser, Hwangbo, Hoepflinger, et~al.]{gehring2016practice}
C.~Gehring, S.~Coros, M.~Hutter, C.~D. Bellicoso, H.~Heijnen, R.~Diethelm, M.~Bloesch, P.~Fankhauser, J.~Hwangbo, M.~Hoepflinger \emph{et~al.}, ``Practice makes perfect: An optimization-based approach to controlling agile motions for a quadruped robot,'' \emph{IEEE Robotics \& Automation Magazine}, vol.~23, no.~1, pp. 34--43, 2016.

\bibitem[Song et~al.(2022)Song, Yue, Sun, Ling, Wei, Gui, and Liu]{song2022optimal}
Z.~Song, L.~Yue, G.~Sun, Y.~Ling, H.~Wei, L.~Gui, and Y.-H. Liu, ``An optimal motion planning framework for quadruped jumping,'' in \emph{2022 IEEE/RSJ International Conference on Intelligent Robots and Systems (IROS)}.\hskip 1em plus 0.5em minus 0.4em\relax IEEE, 2022, pp. 11\,366--11\,373.

\bibitem[Pandala et~al.(2022)Pandala, Fawcett, Rosolia, Ames, and Hamed]{pandala2022robust}
A.~Pandala, R.~T. Fawcett, U.~Rosolia, A.~D. Ames, and K.~A. Hamed, ``Robust predictive control for quadrupedal locomotion: Learning to close the gap between reduced-and full-order models,'' \emph{IEEE Robotics and Automation Letters}, vol.~7, no.~3, pp. 6622--6629, 2022.

\bibitem[Li and Wensing(2024)]{li2024cafe}
H.~Li and P.~M. Wensing, ``Cafe-mpc: A cascaded-fidelity model predictive control framework with tuning-free whole-body control,'' \emph{arXiv preprint arXiv:2403.03995}, 2024.

\bibitem[Grandia et~al.(2019)Grandia, Farshidian, Ranftl, and Hutter]{grandia2019feedback}
R.~Grandia, F.~Farshidian, R.~Ranftl, and M.~Hutter, ``Feedback mpc for torque-controlled legged robots,'' in \emph{2019 IEEE/RSJ International Conference on Intelligent Robots and Systems (IROS)}, 2019, pp. 4730--4737.

\bibitem[Qi et~al.(2021)Qi, Lin, Hong, Chen, and Zhang]{qi2021perceptive}
S.~Qi, W.~Lin, Z.~Hong, H.~Chen, and W.~Zhang, ``Perceptive autonomous stair climbing for quadrupedal robots,'' in \emph{2021 IEEE/RSJ International Conference on Intelligent Robots and Systems (IROS)}.\hskip 1em plus 0.5em minus 0.4em\relax IEEE, 2021, pp. 2313--2320.

\bibitem[Park et~al.(2021)Park, Wensing, and Kim]{park2021jumping}
H.-W. Park, P.~M. Wensing, and S.~Kim, ``Jumping over obstacles with mit cheetah 2,'' \emph{Robotics and Autonomous Systems}, vol. 136, p. 103703, 2021.

\bibitem[Grandia et~al.(2023)Grandia, Jenelten, Yang, Farshidian, and Hutter]{grandia2023perceptive}
R.~Grandia, F.~Jenelten, S.~Yang, F.~Farshidian, and M.~Hutter, ``Perceptive locomotion through nonlinear model-predictive control,'' \emph{IEEE Transactions on Robotics}, vol.~39, no.~5, pp. 3402--3421, 2023.

\bibitem[Margolis et~al.(2022)Margolis, Yang, Paigwar, Chen, and Agrawal]{margolis2022rapid}
G.~B. Margolis, G.~Yang, K.~Paigwar, T.~Chen, and P.~Agrawal, ``Rapid locomotion via reinforcement learning,'' \emph{arXiv preprint arXiv:2205.02824}, 2022.

\bibitem[Margolis and Agrawal(2023)]{walk_these_ways}
G.~B. Margolis and P.~Agrawal, ``Walk these ways: Tuning robot control for generalization with multiplicity of behavior,'' in \emph{Conference on Robot Learning}.\hskip 1em plus 0.5em minus 0.4em\relax PMLR, 2023, pp. 22--31.

\bibitem[Agarwal et~al.(2022)Agarwal, Kumar, Malik, and Pathak]{rma_vision}
A.~Agarwal, A.~Kumar, J.~Malik, and D.~Pathak, ``Legged locomotion in challenging terrains using egocentric vision,'' \emph{arXiv preprint arXiv:2211.07638}, 2022.

\bibitem[Smith et~al.(2023)Smith, Kew, Li, Luu, Peng, Ha, Tan, and Levine]{twirl}
L.~Smith, J.~C. Kew, T.~Li, L.~Luu, X.~B. Peng, S.~Ha, J.~Tan, and S.~Levine, ``Learning and adapting agile locomotion skills by transferring experience,'' \emph{arXiv preprint arXiv:2304.09834}, 2023.

\bibitem[He et~al.(2024)He, Zhang, Xiao, He, Liu, and Shi]{He-RSS-24}
T.~He, C.~Zhang, W.~Xiao, G.~He, C.~Liu, and G.~Shi, ``{Agile But Safe: Learning Collision-Free High-Speed Legged Locomotion},'' in \emph{Proceedings of Robotics: Science and Systems}, Delft, Netherlands, July 2024.

\bibitem[Makoviychuk et~al.(2021)Makoviychuk, Wawrzyniak, Guo, Lu, Storey, Macklin, Hoeller, Rudin, Allshire, Handa, and State]{makoviychuk2021isaac}
V.~Makoviychuk, L.~Wawrzyniak, Y.~Guo, M.~Lu, K.~Storey, M.~Macklin, D.~Hoeller, N.~Rudin, A.~Allshire, A.~Handa, and G.~State, ``Isaac gym: High performance gpu-based physics simulation for robot learning,'' 2021.

\bibitem[Miki et~al.(2022)Miki, Lee, Hwangbo, Wellhausen, Koltun, and Hutter]{eth_hike}
T.~Miki, J.~Lee, J.~Hwangbo, L.~Wellhausen, V.~Koltun, and M.~Hutter, ``Learning robust perceptive locomotion for quadrupedal robots in the wild,'' \emph{Science Robotics}, vol.~7, no.~62, p. eabk2822, 2022.

\bibitem[Xie et~al.(2020)Xie, Ling, Kim, and van~de Panne]{xie2020allsteps}
Z.~Xie, H.~Y. Ling, N.~H. Kim, and M.~van~de Panne, ``Allsteps: curriculum-driven learning of stepping stone skills,'' in \emph{Computer Graphics Forum}, vol.~39, no.~8.\hskip 1em plus 0.5em minus 0.4em\relax Wiley Online Library, 2020, pp. 213--224.

\bibitem[Yu et~al.(2021)Yu, Jain, Escontrela, Iscen, Xu, Coumans, Ha, Tan, and Zhang]{googlevisuallocomotion}
W.~Yu, D.~Jain, A.~Escontrela, A.~Iscen, P.~Xu, E.~Coumans, S.~Ha, J.~Tan, and T.~Zhang, ``Visual-locomotion: Learning to walk on complex terrains with vision,'' in \emph{5th Annual Conference on Robot Learning}, 2021.

\bibitem[Da et~al.(2020)Da, Xie, Hoeller, Boots, Anandkumar, Zhu, Babich, and Garg]{laikagonvidia}
X.~Da, Z.~Xie, D.~Hoeller, B.~Boots, A.~Anandkumar, Y.~Zhu, B.~Babich, and A.~Garg, ``Learning a contact-adaptive controller for robust, efficient legged locomotion,'' \emph{arXiv preprint arXiv:2009.10019}, 2020.

\bibitem[Xie et~al.(2021)Xie, Da, Babich, Garg, and van~de Panne]{glide}
Z.~Xie, X.~Da, B.~Babich, A.~Garg, and M.~van~de Panne, ``Glide: Generalizable quadrupedal locomotion in diverse environments with a centroidal model,'' \emph{arXiv preprint arXiv:2104.09771}, 2021.

\bibitem[Yang et~al.(2022)Yang, Zhang, Coumans, Tan, and Boots]{fast_and_efficient}
Y.~Yang, T.~Zhang, E.~Coumans, J.~Tan, and B.~Boots, ``Fast and efficient locomotion via learned gait transitions,'' in \emph{Conference on Robot Learning}.\hskip 1em plus 0.5em minus 0.4em\relax PMLR, 2022, pp. 773--783.

\bibitem[Yang et~al.(2023{\natexlab{b}})Yang, Meng, Yu, Zhang, Tan, and Boots]{yang2023continuous}
Y.~Yang, X.~Meng, W.~Yu, T.~Zhang, J.~Tan, and B.~Boots, ``Continuous versatile jumping using learned action residuals,'' \emph{arXiv preprint arXiv:2304.08663}, 2023.

\bibitem[Bellegarda and Nguyen(2020)]{bellegarda2020robust}
G.~Bellegarda and Q.~Nguyen, ``Robust quadruped jumping via deep reinforcement learning,'' \emph{arXiv preprint arXiv:2011.07089}, 2020.

\bibitem[Ross et~al.(2011)Ross, Gordon, and Bagnell]{dagger}
S.~Ross, G.~Gordon, and D.~Bagnell, ``A reduction of imitation learning and structured prediction to no-regret online learning,'' in \emph{Proceedings of the fourteenth international conference on artificial intelligence and statistics}.\hskip 1em plus 0.5em minus 0.4em\relax JMLR Workshop and Conference Proceedings, 2011, pp. 627--635.

\bibitem[Rudin et~al.(2022)Rudin, Hoeller, Reist, and Hutter]{rudin2022learning}
N.~Rudin, D.~Hoeller, P.~Reist, and M.~Hutter, ``Learning to walk in minutes using massively parallel deep reinforcement learning,'' in \emph{Conference on Robot Learning}.\hskip 1em plus 0.5em minus 0.4em\relax PMLR, 2022, pp. 91--100.

\bibitem[Raibert(1986)]{raibertcontroller}
M.~H. Raibert, \emph{Legged robots that balance}.\hskip 1em plus 0.5em minus 0.4em\relax MIT press, 1986.

\bibitem[Paszke et~al.(2019)Paszke, Gross, Massa, Lerer, Bradbury, Chanan, Killeen, Lin, Gimelshein, Antiga, et~al.]{pytorch}
A.~Paszke, S.~Gross, F.~Massa, A.~Lerer, J.~Bradbury, G.~Chanan, T.~Killeen, Z.~Lin, N.~Gimelshein, L.~Antiga \emph{et~al.}, ``Pytorch: An imperative style, high-performance deep learning library,'' \emph{Advances in neural information processing systems}, vol.~32, 2019.

\bibitem[Schulman et~al.(2017)Schulman, Wolski, Dhariwal, Radford, and Klimov]{ppo}
J.~Schulman, F.~Wolski, P.~Dhariwal, A.~Radford, and O.~Klimov, ``Proximal policy optimization algorithms,'' \emph{arXiv preprint arXiv:1707.06347}, 2017.

\end{thebibliography}
\end{document}